\documentclass[conference]{IEEEtran}

\IEEEoverridecommandlockouts % Required for \thanks and \IEEEaftertitletext

\ifdefined\pdfminorversion
\fi
\usepackage{cite}
\usepackage[hidelinks]{hyperref}
\usepackage{amsmath,amssymb,amsfonts}
\usepackage{graphicx}
\usepackage{cleveref}
\usepackage{capt-of} % Provides \captionof without replacing IEEE captions
\usepackage{hyperref}
\hypersetup{
    colorlinks = true}
\usepackage{textcomp}
\usepackage{xcolor}
\usepackage{booktabs}
\usepackage{pgfplots}
\usepackage{tikz}
\usetikzlibrary{matrix}
\usepackage{url}
\usepackage{tabularx}

\colorlet{ObservedCell}{green!30}
\colorlet{HeldOutCell}{orange!30}

\tikzset{
    cell/.style={minimum width=2cm, minimum height=1cm, anchor=center}
}

\pgfplotsset{compat=1.18}
\usetikzlibrary{matrix,positioning}

\title{Spa-Bench: A Real-Robot Benchmark for Spatial Grounding
in Vision-Language-Action Models}

\author{
\IEEEauthorblockN{
Justin Tien-Smith$^*$ \quad
Matthew Pidden$^*$ \quad
Rolandos Alexandros Potamias$^\dagger$
}
\IEEEauthorblockA{
Imperial College London\\
{\url{https://justintiensmith.github.io/Spa-Bench/}}
}
\thanks{$^*$Equal contribution. $^\dagger$Advising author.}
}

\IEEEaftertitletext{%
\begin{minipage}{\textwidth}

{\centering
\includegraphics[
    width=\textwidth,
    height=0.40\textheight,
    keepaspectratio
]{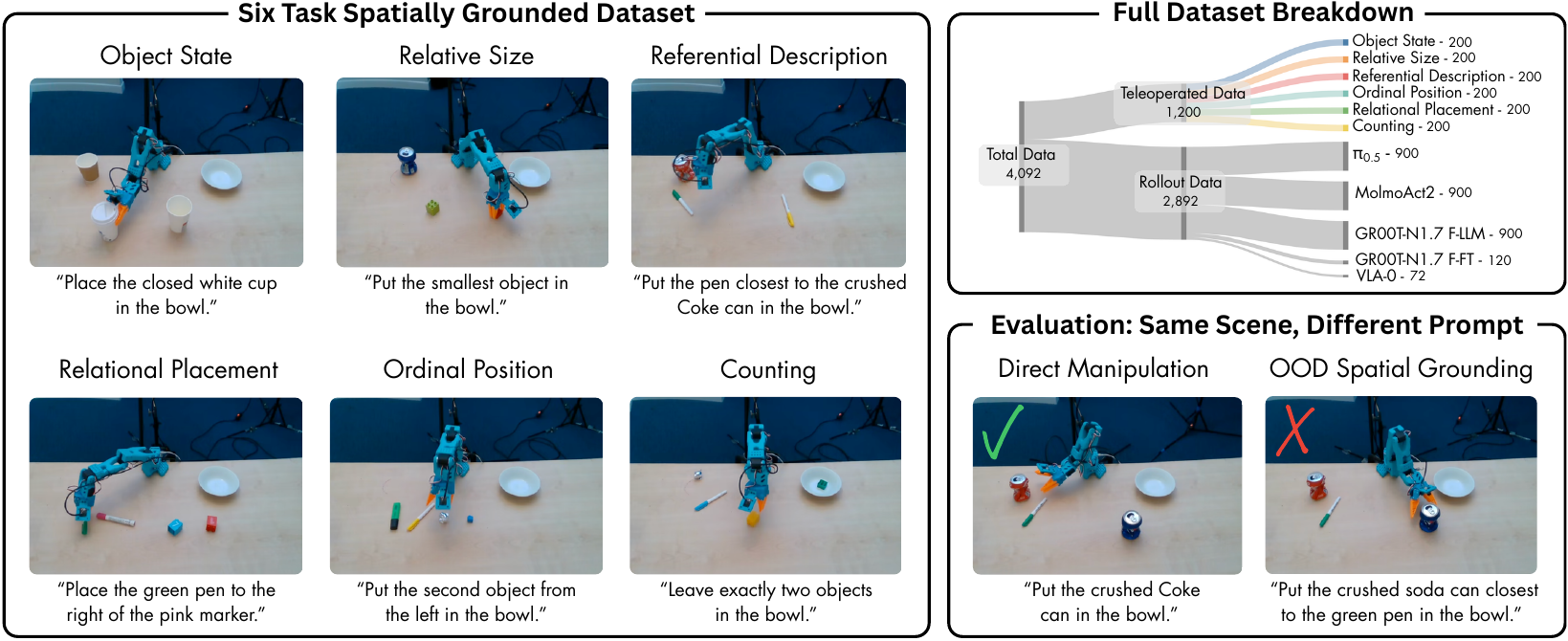}
\par}

\captionof{figure}{
\textbf{Overview of Spa-Bench.}
Left: representative examples from the six spatially grounded tasks: \emph{Object State}, \emph{Relative Size}, \emph{Referential Description}, \emph{Relational Placement}, \emph{Ordinal Position}, and \emph{Counting}.
Top right: composition of the 4,092-episode dataset, comprising 1,200 human-teleoperated demonstrations (200 per task family) and 2,892 real-robot evaluation rollouts across five VLA policies. Bottom right: an illustrative matched-scene comparison in which the scene and required physical action remain the same while a direct manipulation instruction is replaced by an OOD spatially grounded instruction. This comparison helps distinguish spatial-grounding failures from an inability to execute the underlying manipulation.}
\label{fig:teaser}
\end{minipage}
\vspace{0.5\baselineskip}
}

\begin{document}

\maketitle

\begin{abstract}
Vision-language-action models (VLAs) offer a promising route toward general-purpose robotics by transferring semantic knowledge from large-scale vision-language pretraining into language-conditioned control. However, most VLA evaluations emphasize aggregate manipulation success under changes in objects, instructions, or environments, leaving unclear how reliably these policies ground spatial descriptions in the physical scenes where they act. We therefore ask a sharper question: \textit{how reliably can current VLAs translate spatial descriptions into appropriate real-robot behavior?} To investigate this, we introduce Spa-Bench, a real-robot benchmark that evaluates spatial grounding across six tabletop tasks covering \textit{Object State}, \textit{Relative Size}, \textit{Referential Description}, \textit{Relational Placement}, \textit{Ordinal Position}, and \textit{Counting}. Our primary evaluation compares in-distribution (ID) and out-of-distribution (OOD) spatial configurations while retaining training support for the relevant objects, concepts, and actions, reducing object and motor unfamiliarity as alternative explanations for failure. Four diagnostic conditions further probe physical execution, sensitivity to wording, object-task associations, and recognition that no action is required. We release 4,092 episodes collected on the SO-101 robot arm: 1,200 human-teleoperated demonstrations and 2,892 real-robot evaluation rollouts from five state-of-the-art VLA policies. Across the three fully evaluated policies, success falls from 67.5--73.3\% on ID spatial configurations to 40.7--52.0\% on OOD spatial configurations, while matched direct instructions improve success by 28.3--40.0 percentage points. These results show that the evaluated VLAs can often execute the required manipulations but lack robust spatial grounding. Project materials, including videos, code, and benchmark data, are available at \url{https://justintiensmith.github.io/Spa-Bench/}.

\end{abstract}


\section{Introduction}

Vision-language-action models (VLAs) have emerged as a promising approach to general-purpose robot learning. By connecting pretrained vision-language representations to robot actions, these models can transfer semantic knowledge acquired from internet-scale data into language-conditioned control \cite{brohan2023rt2,black2025pi05,nvidia2025gr00t,fang2026molmoact2}. This transfer promises robots that can respond to new objects, instructions, and environments without requiring a separately trained policy for every behavior.

Policies evaluated on existing benchmarks have demonstrated increasingly broad manipulation capabilities, including generalization across objects, instructions, visual conditions, and task sequences \cite{zhang2024vlabench,zhou2025liberopro,gao2026stargen}. Nevertheless, completing a manipulation task does not by itself establish that a policy has correctly interpreted the spatial structure of its environment. Spatial understanding in existing benchmarks is typically embedded within an overall task-success metric and is consequently entangled with object recognition, language interpretation, and motor execution. If a policy fails to place a green pen to the left of a red block, for example, the failure could arise from incorrect spatial grounding, failure to identify the objects, an unsuccessful grasp, or an inability to execute the required placement due to an inadequate number of demonstrations spanning the full workspace. Furthermore, benchmarks that introduce a fixed scene to mitigate some of these issues introduce a different problem: Success may result from memorizing recurring object locations or action patterns rather than from grounding the instructed spatial relation.

In this work, we use \emph{spatial grounding} to describe the ability to map language about properties, positions, quantities, and relations in the current physical scene to an appropriate target, goal configuration, or decision not to act. We study this capability through the following question:

\begin{quote}
\emph{How reliably can current VLAs ground spatial descriptions into appropriate actions in the real world?}
\end{quote}

Answering this question requires more than measuring success on an ordinary held-out task. A test task that simultaneously introduces an unfamiliar object, instruction, spatial concept, or physical behavior cannot reveal which of these changes caused a failure. A diagnostic spatial benchmark should instead establish familiarity with the constituent objects, concepts, and manipulation behaviors before evaluating whether the policy can use the relevant spatial information in a different physical configuration.

We introduce the \emph{Spatial Grounding Benchmark} (Spa-Bench), a controlled real-robot benchmark designed around this principle. Spa-Bench comprises six tabletop tasks evaluating the grounding of \textit{Object State}, \textit{Relative Size}, \textit{Relational Placement}, \textit{Referential Description}, \textit{Ordinal Position}, and \textit{Counting}. All tasks use the same robot embodiment, workspace, receptacle, and broad pick-and-place action family. The principal source of variation is therefore the scene-grounded rule that determines which object should be manipulated, where it should be placed, or whether movement is even required.

Spa-Bench evaluates policies in ID and OOD spatial configurations. ID configurations measure whether a policy has acquired the spatially grounded behaviors represented in its demonstrations. OOD configurations change the association between a spatial concept and its task argument while preserving training support for the relevant objects, physical actions, and spatial concepts. We implement this controlled split through a \textit{withheld-cell} design: each component of an OOD evaluation configuration appears elsewhere in training, but their particular combination is reserved for evaluation (see Fig.~\ref{fig:withheld_cell_construction}). The purpose of this construction is to reduce object and motor unfamiliarity as alternative explanations for failure.

To characterize spatially grounded performance beyond the primary ID--OOD comparison, Spa-Bench includes four additional diagnostic conditions that probe complementary aspects of policy behavior. Direct-manipulation controls remove the need for spatial inference while preserving the requested robot motion. Prompt-paraphrase trials restate both OOD spatial and direct-manipulation instructions on the same scenes, testing whether their performance gap depends on exact wording. Cross-task-object rollouts vary the usual association between an object and task. Finally, no-intervention scenes test whether a policy can recognize when no action is required.

Spa-Bench contains 1,200 human-teleoperated demonstrations, with 200 demonstrations for each task. We evaluate five fine-tuned state-of-the-art policies that are drawn from four different VLA architectures: VLA-0, MolmoAct2, $\pi_{0.5}$, GR00T-N1.7 Frozen LLM (GR00T-N1.7 F-LLM), and GR00T-N1.7 Full Fine-Tune (GR00T-N1.7 F-FT). Three policies complete the full 900-rollout evaluation protocol, producing 2,700 evaluation episodes, while two partial baselines contribute 192 episodes. In total, we release 4,092 SO-101 episodes, including the demonstrations, rollout videos, task specifications, and documented scene layouts.

Across the three fully evaluated policies, ID spatial-configuration success ranges from 67.5\% to 73.3\%, but success decreases to 40.7\%--52.0\% in OOD spatial configurations. On the OOD spatial configuration matched scene subset, direct-manipulation instructions achieve 81.7\%--89.2\% success, showing that many of the physical behaviors are executable even when the policy fails under a spatially grounded instruction. Performance is strongly task dependent: all three policies fail every OOD \textit{Counting} trial, almost always without acting; other tasks are dominated by target-selection or spatial-outcome errors. The direct manipulation advantage also persists when both instruction types are paraphrased, revealing that the state-of-the-art VLAs we evaluate do not exhibit robust spatial grounding. In summary, our core contributions are:

\begin{itemize}
    \item We introduce Spa-Bench, a six-task real-robot benchmark for evaluating spatial grounding across ID and OOD physical configurations.

    \item We develop a controlled evaluation protocol that combines a \textit{withheld-cell} split with four diagnostics that probe physical execution, wording sensitivity, object context, and action gating.

    \item We evaluate five fine-tuned VLA policies and release 4,092 SO-101 episodes, including 1,200 human-teleoperated demonstrations and 2,892 evaluation rollouts, together with task specifications and scene layouts.
\end{itemize}

\section{Related Work}

\subsection{Vision-Language-Action Models}

Vision-language-action models (VLAs) connect pretrained multimodal representations to robot actions with the goal of transferring semantic and visual knowledge into closed-loop control. RT-2 helped establish this formulation by representing actions as tokens and co-fine-tuning a vision-language model on web and robot data, demonstrating transfer to novel objects, instructions, and semantically specified behaviors \cite{brohan2023rt2}. OpenVLA subsequently introduced an open-source 7-billion-parameter VLA trained on large-scale robot data and designed for adaptation to downstream embodiments \cite{kim2024openvla}. In parallel, $\pi_0$ coupled a pretrained vision-language backbone to a flow-matching action expert for generating continuous action chunks across multiple embodiments \cite{black2025pi0}.

More recent VLAs have targeted open-world generalization, heterogeneous robot data, embodied reasoning, and practical deployment. $\pi_{0.5}$ extends $\pi_0$ through heterogeneous co-training on robot trajectories, verbal instructions, semantic predictions, and web data \cite{black2025pi05}. The GR00T-N1.7 family uses a dual-system architecture in which a vision-language module interprets the scene and instruction while a diffusion transformer produces continuous actions \cite{nvidia2025gr00t}. VLA-0 investigates whether continuous actions can instead be generated directly as numerical text without introducing a specialized action vocabulary or action head \cite{goyal2025vla0}. MolmoAct2 connects an embodied-reasoning vision-language model to a flow-matching continuous-action expert \cite{fang2026molmoact2}. Because these models differ in pretraining data, training objectives, action representations, and fine-tuning procedures, Spa-Bench compares complete policy systems rather than isolating individual architectural choices. Each fine-tuned policy is evaluated under the same controlled spatial configurations and diagnostic conditions to characterize its spatial generalization.

\subsection{Simulation and Reasoning Benchmarks}

Simulation benchmarks have enabled scalable evaluation of language-conditioned robot behavior. RLBench provides a broad collection of vision-guided manipulation tasks and automatically generated demonstrations \cite{james2019rlbench}, while ALFRED evaluates grounded instruction following over compositional household activities \cite{shridhar2019alfred}. LIBERO studies declarative and procedural knowledge transfer across lifelong-learning task suites \cite{liu2023libero}. RoboCasa expands language-conditioned manipulation to diverse household activities in simulated kitchens \cite{nasiriany2024robocasa}, and BEHAVIOR-1K targets long-horizon everyday activities involving physical and semantic state changes \cite{li2024behavior1k}. VLABench further evaluates semantic knowledge, spatial relationships, physical understanding, and long-horizon reasoning across numerous tasks and objects \cite{zhang2024vlabench}. VIMA-Bench more explicitly separates forms of generalization, including a combinatorial split that evaluates unseen object--texture pairings whose individual components appear during training \cite{jiang2023vima}. These benchmarks establish task breadth and scalable evaluation, but spatial grounding is generally one component of overall task success and therefore remains entangled with perception, planning, and execution.

A complementary line of work examines whether policy performance persists under controlled distribution shifts. The COLOSSEUM evaluates manipulation policies under systematic visual and physical perturbations \cite{pumacay2024colosseum}, while LIBERO-PRO varies manipulated objects, initial states, instructions, and environments, revealing that strong standard-benchmark performance can depend on memorized layouts and action sequences \cite{zhou2025liberopro}. STAR-Gen organizes generalization along visual, semantic, and behavioral axes \cite{gao2026stargen}. BeTTER applies targeted interventions, including spatial-layout shifts, together with kinematic isolation intended to separate reasoning failures from execution limitations \cite{xu2026better}. Other benchmarks evaluate reasoning without requiring closed-loop physical execution. ManipBench probes low-level manipulation reasoning in vision-language models \cite{zhao2025manipbench}, whereas MultihopSpatial evaluates compositional spatial inference and visual grounding over multi-hop relationships \cite{lee2026multihopspatial}. These evaluations provide fine-grained measures of reasoning, but correct visual or textual answers do not necessarily establish that the inferred relation can guide real-robot behavior.

\subsection{Real-World Robot Evaluation}

Real-world benchmarks address the complementary challenges of physical validity, reproducibility, and evaluation at scale. REALM evaluates manipulation policies under controlled perturbations in simulation and validates the resulting trends through physical experiments \cite{sedlacek2025realm}. RoboChallenge provides infrastructure for evaluating embodied policies across multiple real robots and manipulation tasks \cite{yakefu2025robochallenge}. RADAR focuses on real-world dynamics and spatial-physical intelligence while using automated three-dimensional success metrics \cite{chen2026radar}. ManipArena combines diverse real-robot tasks with task-specific OOD conditions, including unseen objects and held-out spatial, ordering, or procedural configurations of familiar objects \cite{sun2026maniparena}. VLA-REPLICA instead emphasizes low-cost cross-laboratory reproducibility through standardized hardware and evaluation on the SO-101 platform \cite{huang2026vlareplica}.

Spa-Bench complements these efforts by isolating spatial grounding under closed-loop real-robot execution. It fixes the robot, workspace, object inventory, and broad manipulation behavior, and places the held-out combination specifically in the mapping between spatial concepts and task arguments. Thus, its OOD configurations preserve training support for the constituent objects, spatial concepts, and physical behaviors. Direct-manipulation controls test whether the relevant manipulation behavior remains executable when the task-specific spatial inference is removed. Prompt paraphrases test whether the performance difference between matched OOD spatial and direct instructions persists under alternative phrasings. Cross-task-object evaluations test sensitivity to familiar object--task associations, while no-intervention scenes test whether policies can correctly withhold action. Together, these diagnostics make Spa-Bench a targeted evaluation of how reliably current VLAs translate spatial descriptions into appropriate behavior in physical scenes. 

\section{Spa-Bench Design}
\label{sec:spa_bench_design}

\subsection{Task Suite}

Spa-Bench evaluates whether a robot policy can ground spatial descriptions in the current physical scene and translate them into an appropriate action. Specifically, the benchmark contains six tabletop tasks spanning three axes of spatial grounding: attribute, relation, and scene structure. The \textit{Object State} and \textit{Relative Size} tasks require selecting an object according to a visually grounded attribute, the \textit{Relational Placement} and \textit{Referential Description} tasks require interpreting directed or object-relative spatial relations whereas the \textit{Ordinal Position} and \textit{Counting} tasks require reasoning about the organization of the scene as a whole.

All six tasks use the same SO-101 robot embodiment, workspace, receptacle, and broad pick-and-place action family. The primary source of variation is therefore the scene-grounded rule that determines which object should be manipulated, where it should be placed, or whether an action is necessary. Five tasks additionally include diagnostic scenes in which no intervention is required because the requested condition already holds or no valid target exists; \textit{Relative Size} is excluded because each comparison set contains a unique smallest and largest object; introducing a size tie would create target ambiguity rather than a clear no-action condition.

All directional and ordinal descriptions are defined from the robot's perspective. For example, \emph{left}, \emph{right}, \emph{front}, and \emph{behind} refer to directions in the robot's workspace rather than directions relative to an individual camera view. Table~\ref{tab:task_overview} summarizes the six tasks and their success criteria.

\begin{table*}[t]
\caption{Spa-Bench tasks. All tasks use the same robot, workspace, receptacle, and broad pick-and-place behavior while varying the spatial information needed to determine the correct response.}
\label{tab:task_overview}
\centering
\footnotesize
\renewcommand{\arraystretch}{1.08}
\begin{tabular}{p{2.8cm}p{2.0cm}p{6.3cm}p{5.8cm}}
\toprule
\textbf{Task} &
\textbf{Grounding Axis} &
\textbf{Example Instruction} &
\textbf{Success Criterion} \\
\midrule

Object State &
Attribute &
``Place the \textbf{\textit{closed}} white cup in the bowl.'' &
The same-category object whose object state matches the instruction is placed in the bowl. \\

Relative Size &
Attribute &
``Place the \textbf{\textit{smallest}} object in the bowl.'' &
The object satisfying the requested relative-size description is placed in the bowl. \\

Relational Placement &
Relation &
``Place the green pen to the \textbf{\textit{left of}} the red block.'' &
The target finishes within 3 inches of the reference and within the requested $\pm45^{\circ}$ directional sector. \\

Referential Description &
Relation &
``Place the pencil sharpener \textbf{\textit{furthest}} from the yellow pen in the bowl.'' &
The object satisfying the specified relation to the reference object is placed in the bowl. \\

Ordinal Position &
Scene structure &
``Place the \textbf{\textit{third}} object \textbf{\textit{from the left}} in the bowl.'' &
The object occupying the requested ordinal position is placed in the bowl. \\

Counting &
Scene structure &
``Leave exactly \textbf{\textit{two}} objects in the bowl.'' &
The requested number of objects remains in the bowl at the end of the episode. \\

\bottomrule
\end{tabular}
\end{table*}

\subsection{Dataset and Physical Setup}

\begin{figure}[t]
    \centering
    \includegraphics[width=\columnwidth]
    {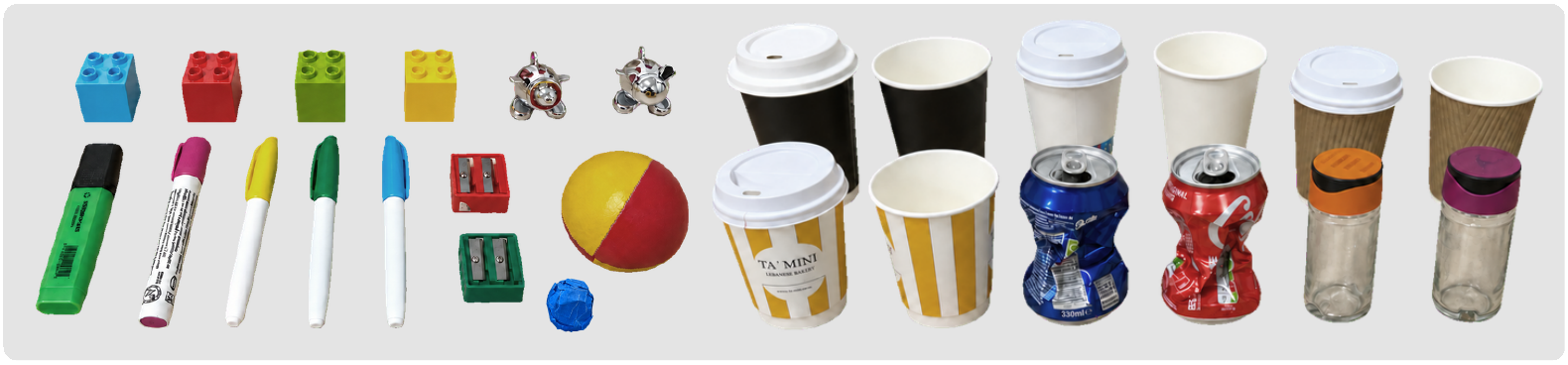}
    \caption{\textbf{Object inventory used in Spa-Bench.} The benchmark uses 27 movable object instances in training and evaluation spanning 11 morphological families. These objects are recombined across the six tasks while the robot embodiment, workspace, and receptacle remain fixed.}
    \label{fig:objects}
\end{figure}

The Spa-Bench dataset contains 1,200 human-teleoperated demonstrations, with 200 demonstrations per task, totaling approximately 5.7 hours and 612,733 recorded frames. Each episode contains synchronized robot states, actions, natural-language instructions, and five RGB camera views. The evaluated policies receive the middle and wrist views. Data collection and evaluation are conducted in the same windowless room with fixed lighting and camera placement. We release the demonstrations, prompt-to-episode checklist, scene specifications, initial robot configurations, and rollout videos. The objects used in our dataset are shown in Fig.~\ref{fig:objects}.

\subsection{Controlled ID and OOD Configurations}
\label{sec:controlled_id_ood}

Existing benchmarks often evaluate policies on held-out tasks that
simultaneously introduce new objects, instructions, spatial concepts, or
physical behaviors. Failure under such a shift cannot be attributed
specifically to spatial grounding. Spa-Bench instead constructs OOD
configurations that preserve training exposure to the relevant objects,
spatial concepts, and manipulation behaviors while withholding selected
pairings between them from spatially grounded instructions.

For each task, we represent the support provided by the spatially grounded
training instructions as a concept--argument matrix. Rows denote spatial concepts and columns denote task-specific arguments, such as object identities, object pairs, or movement types. Pairings that occur in spatially grounded
training instructions form \textit{observed cells}, whereas selected pairings
reserved for OOD evaluation form \textit{withheld cells}. Importantly, a
withheld cell does not imply that its constituent concept, object, or physical
behavior is absent from training. The concept appears with other arguments,
and the evaluation object undergoes the required motion under a direct
instruction that does not reveal the withheld spatial pairing.

\begin{figure}[t]
    \centering
    \includegraphics[width=\columnwidth]
    {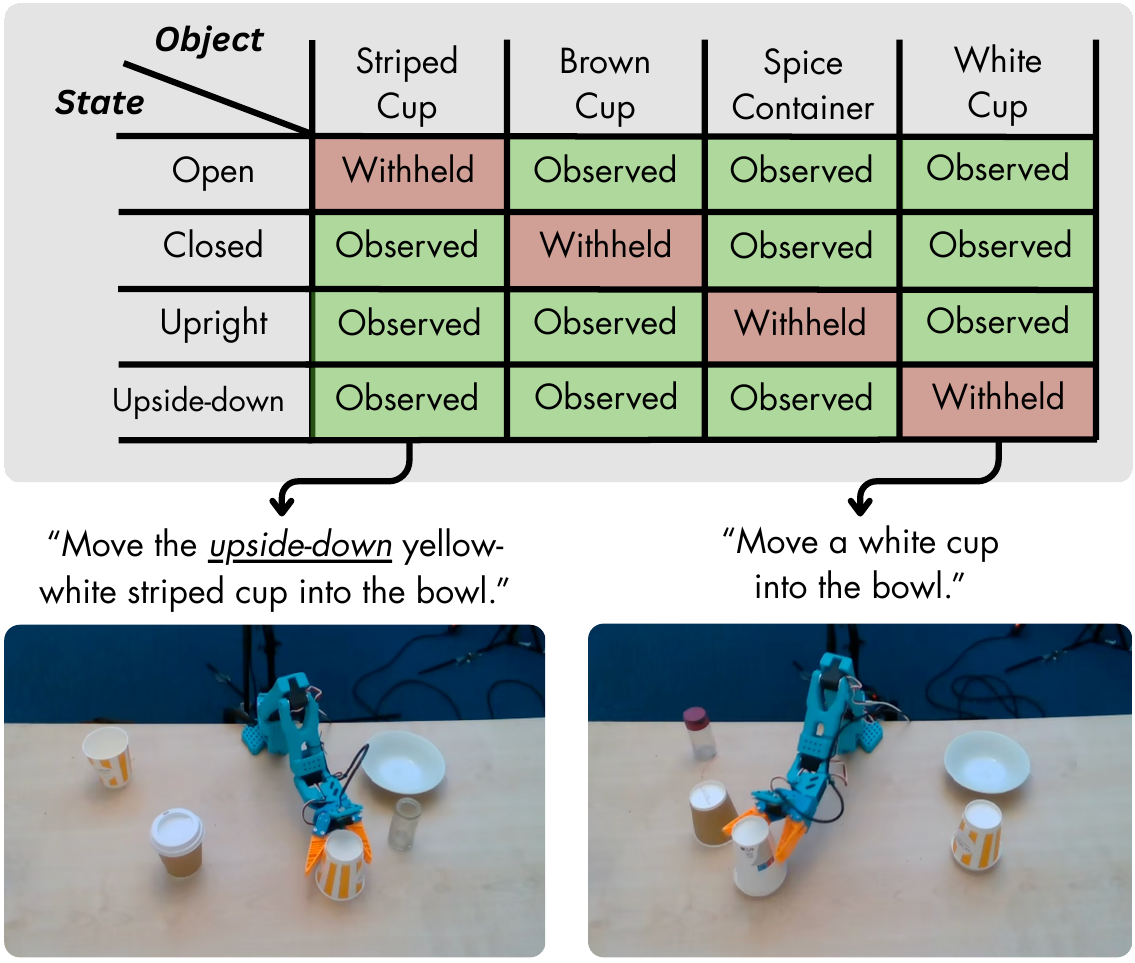}
    \caption{\textbf{Withheld-cell construction for \textit{Object State}.} Rows denote state concepts and columns denote object identities. Observed cells receive state-conditioned instructions during fine-tuning, whereas withheld cells are reserved for OOD evaluation. The concept \textit{upside-down} appears with the striped cup, and the white cup is manipulated under a direct instruction, but the pairing \textit{upside-down white cup} is withheld from spatially grounded training.}
    \label{fig:withheld_cell_construction}
\end{figure}

Figure~\ref{fig:withheld_cell_construction} illustrates this design for the
\textit{Object State} task, where rows denote state concepts and columns denote
object identities. The policy receives spatially grounded instructions
involving \textit{upside-down} objects, including the yellow--white striped
cup. It also sees the upside-down white cup manipulated under the direct
instruction ``Move a white cup into the bowl.'' However, the pairing
\textit{upside-down white cup} never appears in a spatially grounded training
instruction. During OOD evaluation, an upright and an upside-down white cup
appear together, and the policy must select the upside-down instance. Thus,
the object, state concept, and pick-and-place behavior are individually
familiar; only their pairing in spatial language is new. Appendix~\ref{app:complete_matrices}
provides a simplified toy example and the complete matrices for all six tasks.

ID evaluation tests observed cells, measuring performance on spatial pairings
supported during fine-tuning. OOD evaluation tests withheld cells, measuring
whether a policy can recombine familiar spatial concepts, objects, and actions
when their particular pairing was absent from spatially grounded training
instructions.

\subsection{Evaluation Protocol and Conditions}

A rollout is one autonomous policy execution from a specified scene. Each fully evaluated policy completes 900 physical rollouts across the six tasks. Per task, the primary evaluation contains 20 ID and 50 OOD spatial-configuration rollouts. ID configurations evaluate concept--argument pairings present in spatially grounded training instructions (\textit{observed cells}), whereas OOD configurations evaluate pairings absent from those instructions despite training exposure to the constituent concepts, objects, and motions (\textit{withheld cells}).

\begin{figure}[t]
    \centering
    \includegraphics[width=\columnwidth]{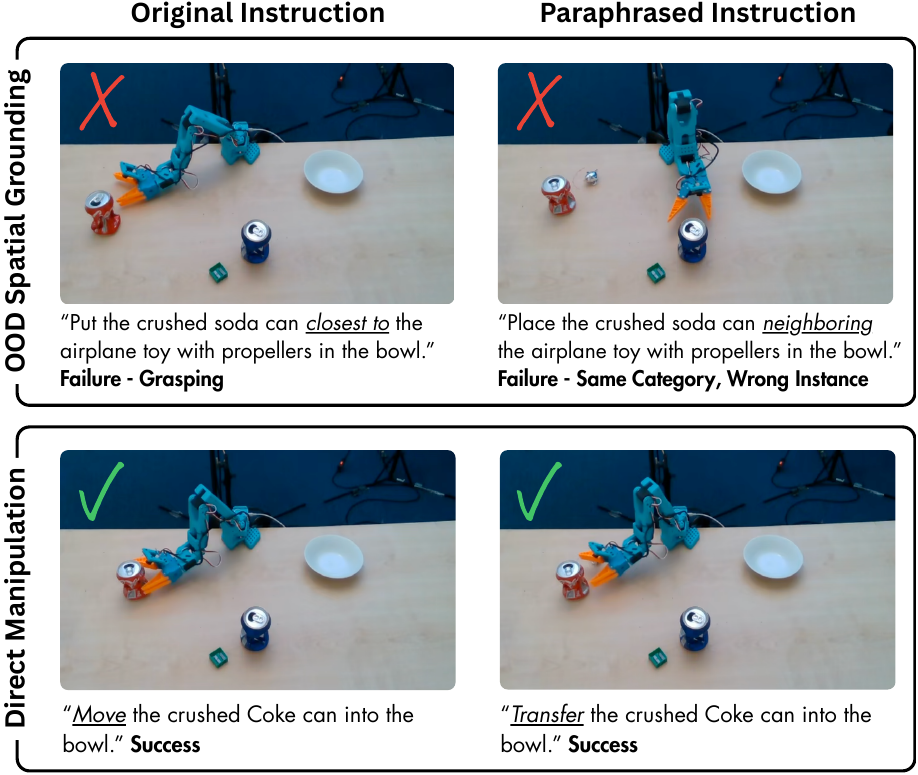}
    \caption{\textbf{Matched-scene diagnostics.} The same scene setup and intended object motion are evaluated under four instruction conditions: OOD spatial, direct manipulation, and a paraphrase of each. In this \textit{Referential Description} example, the direct instruction names the target explicitly, whereas the spatial instruction requires resolving it from the scene.}
    \label{fig:matched_scene_example}
\end{figure}

For 20 of the 50 OOD scene specifications per task, we repeat the evaluation with a direct-manipulation instruction that removes the need for task-specific spatial inference while preserving the scene setup and target object. These control instructions also occur for all objects during fine-tuning, but less frequently than the spatially grounded instructions. For example, the \textit{Referential Description} instruction ``Put the crushed soda can closest to the airplane toy with propellers in the bowl'' requires resolving a relational referent. Its direct control, ``Move the crushed Coke can into the bowl,'' requests the same robot motion but names the target explicitly (Fig.~\ref{fig:matched_scene_example}). A direct-control success paired with a spatial wrong-instance failure therefore provides evidence that referent resolution, rather than pick-and-place execution alone, caused the failure.

For 10 of these 20 matched scenes per task, we also provide paraphrased versions of both instruction types, producing four conditions: original OOD spatial, paraphrased OOD spatial, original direct manipulation, and paraphrased direct manipulation. Within each original--paraphrase pair, the scene setup, intended target, and instruction meaning remain the same while the phrasing changes. This measures wording sensitivity and tests whether the spatial--direct performance difference persists under alternative phrasings (also shown in Fig.~\ref{fig:matched_scene_example}).

An additional 20 rollouts per task evaluate whether ID spatial grounding transfers to objects normally associated with another task. These cross-task-object rollouts retain a familiar spatial concept and manipulation behavior while changing the object context. Because no object from another task instantiates all required \textit{Object State} conditions, that task instead uses a separate novel-object pilot and is excluded from the strict cross-task-object aggregate.

Five tasks additionally include 20 no-intervention rollouts. In \textit{Counting} and \textit{Relational Placement}, the requested count or relation already holds. In \textit{Object State}, \textit{Referential Description}, and \textit{Ordinal Position}, the instruction has no valid target in the scene. Success requires initiating no arm or gripper motion. \textit{Relative Size} omits this condition because every comparison set contains a unique smallest and largest object. \textit{Relational Placement} also includes 20 matched movement-required controls to distinguish recognition that a relation already holds from a general tendency to remain stationary.

These task-specific adjustments produce 170 rollouts for \textit{Relational Placement}, 130 for \textit{Relative Size}, and 150 for each remaining task, totaling 900 rollouts per fully evaluated policy.

\subsection{Cross-Policy Scene Matching, Scoring, and Analysis}

To support comparisons across policies, corresponding rollouts are manually reset to the same scene specification (e.g., evaluation 103 for MolmoAct2 uses the layout specified for evaluation 103 for $\pi_{0.5}$). Resets reproduce object identities and locations using reference scene images; pixel coordinates are not matched. Within each task, the 20 direct-manipulation rollouts reuse a subset of the 50 OOD scene specifications, and 10 of the direct-manipulation scene specifications are additionally reused for both paraphrase conditions.

The primary outcome is binary physical task success defined in Table~\ref{tab:task_overview}; partial credit is not assigned. Rollouts are limited to 60 seconds for all models except VLA-0, which receives 90 seconds due to its slower inference rate. No human intervention occurs after the rollout timer and scoring begin. We report success rates with two-sided Wilson 95\% confidence intervals \cite{wilson1927}. Matched-scene differences are reported as descriptive paired contrasts computed only over the corresponding reused scene subsets. Qualitative failure annotations are assigned separately and do not alter binary scoring.

\section{Models and Experimental Setup}
\label{sec:models_setup}

We evaluate five VLAs from four representative architectures: VLA-0 \cite{goyal2025vla0}, MolmoAct2 \cite{fang2026molmoact2}, $\pi_{0.5}$ \cite{black2025pi05}, and two GR00T-N1.7 adaptations \cite{nvidia2025gr00t}. These architectures span textual \cite{goyal2025vla0} and continuous \cite{black2025pi05,nvidia2025gr00t,fang2026molmoact2} action representations, with action-chunk horizons from 8 to 50 steps \cite{goyal2025vla0,black2025pi05,nvidia2025gr00t,fang2026molmoact2}. Because these factors vary across architectures, cross-architecture performance differences cannot be attributed to any single design choice. The two GR00T-N1.7 adaptations provide a narrower within-architecture comparison of fine-tuning strategies. GR00T-N1.7 F-LLM freezes only the language-model parameters; it continues to receive language instructions while its visual, multimodal, and action-generation modules are updated. GR00T-N1.7 F-FT updates all parameters.

\subsection{Training and Deployment Summary}

Each policy receives the six-dimensional robot state alongside middle- and
wrist-camera observations. All policies are trained for 12 epochs with a
global batch size of 96 on the same 1,200 episode identities using their
native preprocessing and fine-tuning procedures. To suppress high-frequency
action jitter, both GR00T-N1.7 variants use a causal action-smoothing filter
during deployment \cite{anr_gr00t_experiments}. Appendix~\ref{app:model_details}
documents the model-specific preprocessing, optimization, deployment filter,
and initialization procedures.

\subsection{Evaluation Scope}

MolmoAct2, $\pi_{0.5}$, and GR00T-N1.7 F-LLM complete all 900 rollouts and support the full ID--OOD and diagnostic analyses. GR00T-N1.7 F-FT completes only the 120-rollout ID condition, while VLA-0 completes 72 ID rollouts; both are reported as partial ID baselines. Across the five policies, the evaluation contains 2,892 real-robot rollouts. Diagnostic conditions are analyzed separately, with the matched-control gap computed as direct-manipulation success minus OOD success on the corresponding 120 matching scenes. Prompt-paraphrase results are separated into paraphrases of OOD spatial instructions and paraphrases of direct manipulation instructions. The strict cross-task-object aggregate excludes the separate 20-trial \textit{Object State} task novel-object pilot. No-intervention performance is reported by task because the reason for not acting differs across tasks.

\section{Experimental Results}
\label{sec:results}

\subsection{Spatial Grounding: ID and OOD Configurations}

\begin{figure*}[t]
    \centering
    \includegraphics[width=\textwidth]{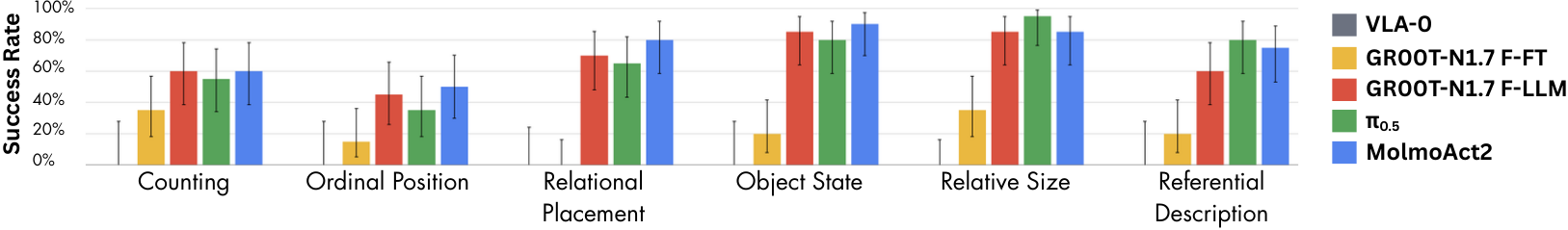}
    \caption{\textbf{Task-level ID spatial-grounding success.}
Bars show per-task success rates, and whiskers indicate two-sided Wilson 95\% confidence intervals. Complete ID evaluations contain 20 trials per task (120 total); VLA-0 contains 72 trials, with at least 10 per task. GR00T-N1.7 F-FT and VLA-0 are reported only as diagnostic ID baselines because their low ID success would make subsequent OOD failures difficult to interpret.}
    \label{fig:familiar_by_task}
\end{figure*}

ID spatial grounding first establishes whether each policy acquired the training-supported spatial behaviors. MolmoAct2 achieved 88/120 successes (73.3\%, Wilson 95\% CI: 64.8--80.4\%), $\pi_{0.5}$ achieved 82/120 (68.3\%, 59.6--76.0\%), and GR00T-N1.7 F-LLM achieved 81/120 (67.5\%, 58.7--75.2\%). Because neither partial baseline established adequate ID task
competence (25/120 for GR00T-N1.7 F-FT and 0/72 for VLA-0), their OOD failures would not distinguish failed recombination from failure to acquire the underlying tasks; both
are therefore retained only as diagnostic ID baselines. Fig.~\ref{fig:familiar_by_task} reports the task-level
ID results for all five policies.

\begin{figure*}[t]
    \centering
    \includegraphics[width=\textwidth]
    {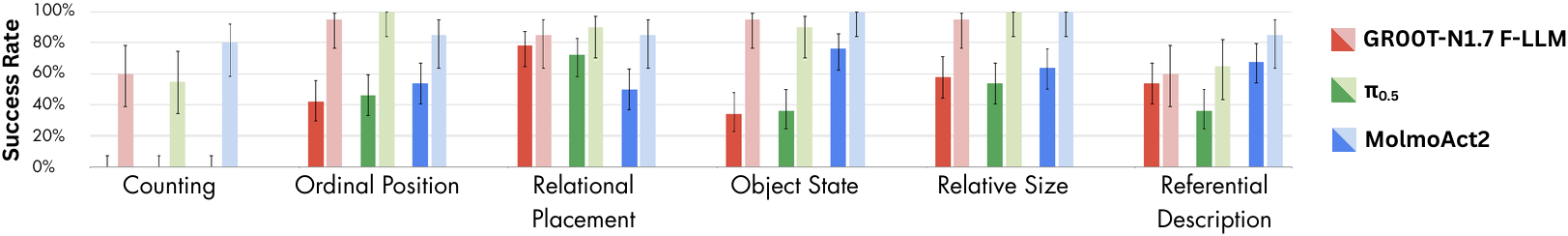}
    \caption{\textbf{Task-level OOD spatial-grounding and
    direct-manipulation success.} Colors denote policies; dark and
    light bars denote OOD spatial and direct-manipulation instructions,
    respectively. Whiskers show two-sided Wilson 95\% confidence
    intervals. OOD bars use 50 trials per task and direct bars use 20;
    the matched comparison in Sec.~\ref{sec:matched_scene_diagnostics}
    uses only the corresponding 20 OOD scenes.}
    \label{fig:novel_and_manipulation_by_task}
\end{figure*}

Performance decreased on the configurations withheld by the
\textit{withheld-cell} split. Across 300 OOD spatial-grounding rollouts per
policy, MolmoAct2 achieved 156/300 successes (52.0\%, Wilson 95\%
CI: 46.4--57.6\%), GR00T-N1.7 F-LLM achieved 133/300
(44.3\%, 38.8--50.0\%), and $\pi_{0.5}$ achieved 122/300
(40.7\%, 35.3--46.3\%). Performance varied by task rather than
producing a consistent model ranking: GR00T-N1.7 F-LLM performed
best on \textit{Relational Placement}, whereas MolmoAct2 performed
best on \textit{Object State} and \textit{Referential Description}.
All three policies failed every OOD \textit{Counting} rollout
(Fig.~\ref{fig:novel_and_manipulation_by_task}), showing that
familiarity with objects and physical transitions
did not make the withheld count-conditioned combinations usable.
Relative to ID, this decrease indicates incomplete recombination of familiar spatial concepts and actions rather than a uniform
loss of task competence.

\subsection{Matched-Scene Diagnostics}
\label{sec:matched_scene_diagnostics}

To determine whether the OOD decrease reflects spatial understanding
rather than robot execution, 20 OOD scene specifications per task
are repeated under direct-manipulation instructions, yielding 120 matched scenes per policy. MolmoAct2 achieved 73/120 successes
(60.8\%) under OOD spatial instructions and 107/120 (89.2\%)
under direct manipulation, a 28.3-percentage-point difference.
The corresponding results were 54/120 (45.0\%) vs. 100/120
(83.3\%) for $\pi_{0.5}$, a 38.3-point difference, and 50/120
(41.7\%) vs. 98/120 (81.7\%) for GR00T-N1.7 F-LLM, a
40.0-point difference. Because these comparisons preserve the scene setup and relevant manipulation behavior while removing
the task-specific spatial concept, the consistently higher
direct-control success shows that execution difficulty alone does
not account for the OOD spatial-grounding failures
(Fig.~\ref{fig:novel_and_manipulation_by_task}).

\begin{figure}[t]
    \centering
    \includegraphics[width=.95\columnwidth]{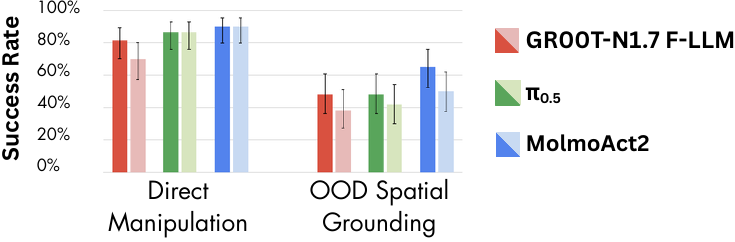}
    \caption{\textbf{Prompt-paraphrase diagnostic on 60 matched scenes.} Dark bars show original instructions; light bars show paraphrased instructions for the same scenes. Whiskers show two-sided Wilson 95\% confidence
    intervals.}
    \label{fig:paraphrasing_comparison}
\end{figure}

To test whether this contrast depends on exact wording, 10 matched
scenes per task also receive paraphrased versions of
both instruction types. On this 60-scene subset, spatial success
decreased from 48.3\% to 38.3\% for GR00T-N1.7 F-LLM, from
48.3\% to 41.7\% for $\pi_{0.5}$, and from 65.0\% to 50.0\%
for MolmoAct2. Under direct-manipulation instructions, success
decreased from 81.7\% to 70.0\% for GR00T-N1.7 F-LLM and
remained at 86.7\% for $\pi_{0.5}$ and 90.0\% for
MolmoAct2. Thus, paraphrasing consistently reduced spatial success,
whereas its effect on direct instructions was model dependent (Fig.~\ref{fig:paraphrasing_comparison}).

\subsection{Cross-Task-Object Transfer}

\begin{figure}[t]
    \centering
    \includegraphics[width=.95\columnwidth]
    {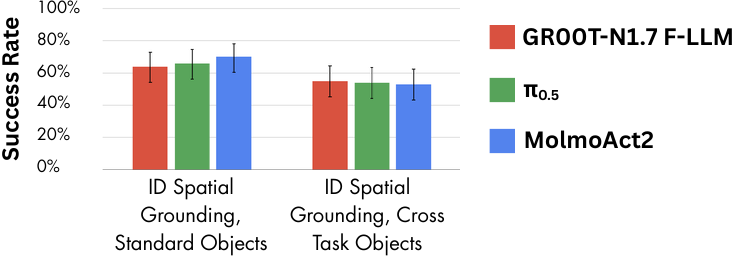}
    \caption{\textbf{Cross-task-object diagnostic.} Bars compare ID spatial-grounding success with standard task objects and with familiar objects transferred from other task families. Each condition contains 100 rollouts across five comparable tasks; whiskers show two-sided Wilson 95\% confidence intervals.}
    \label{fig:cross_task_objects}
\end{figure}

Cross-task-object trials test whether ID spatial grounding transfers to objects from other tasks (Fig.~\ref{fig:cross_task_objects}). Across five comparable tasks, success fell from 64/100 to 55/100 for GR00T-N1.7 F-LLM, from 66/100 to 54/100 for $\pi_{0.5}$, and from 70/100 to 53/100 for MolmoAct2. Because the conditions cover the same tasks but are not paired scenes differing only in object identity, these decreases are descriptive. Their consistent direction nevertheless suggests sensitivity to object--task associations learned during fine-tuning. In the \textit{Relational Placement} trials, none of the three policies defaulted to placing the juggling ball in the bowl, although every fine-tuning demonstration that manipulated the ball placed it in the bowl. This suggests some transfer beyond the learned object--goal association, although the requested relation was not always satisfied.

\subsection{Action Gating and No-Intervention Behavior}

\begin{figure*}[t]
    \centering    
    \includegraphics[width=\textwidth]{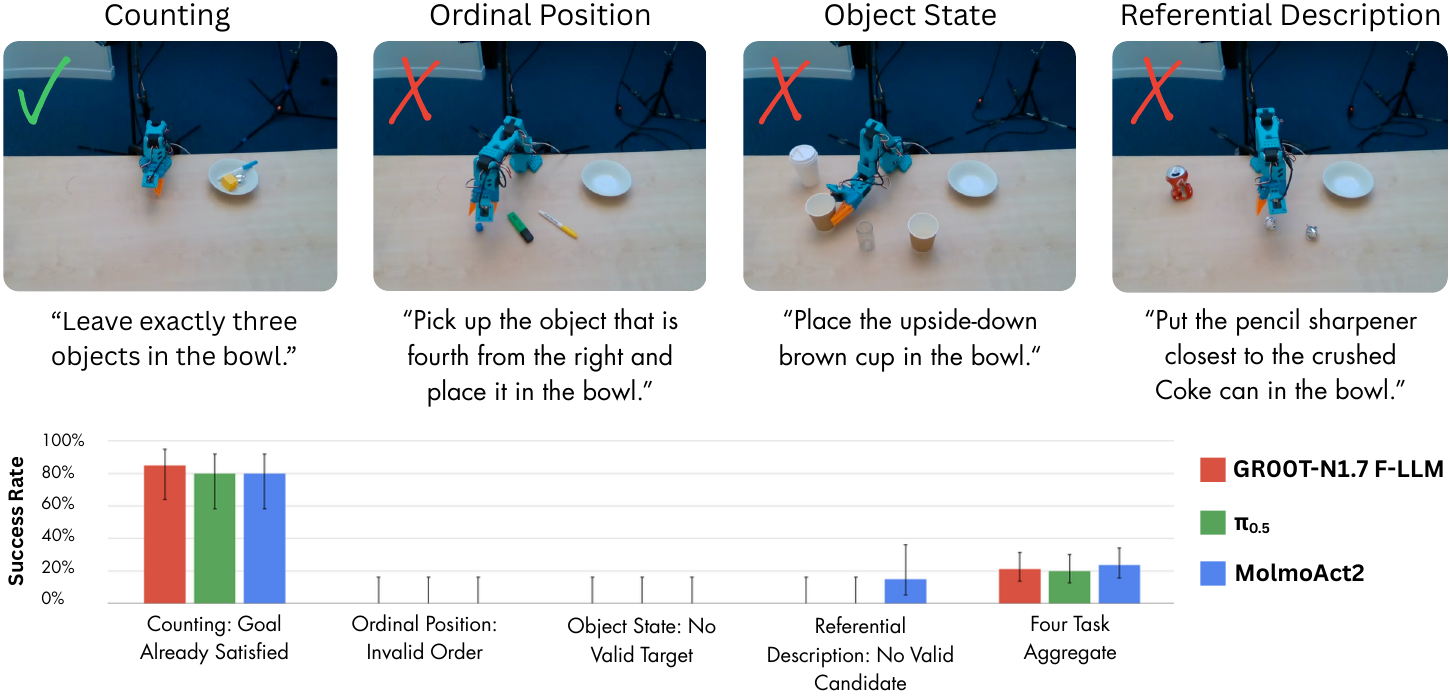}
    \caption{\textbf{No-intervention success in four non-relational diagnostics.} The top row shows representative rollouts; the bars summarize all trials. Success requires withholding task-directed movement when the requested count already holds, the requested ordinal position is invalid, or no object satisfies the requested state or referential description. Each task contains 20 rollouts per policy; the four-task aggregate contains 80. Whiskers show two-sided Wilson 95\% confidence intervals. \textit{Relational Placement} is evaluated separately using matched corrective-movement controls.}
    \label{fig:action_gating_diagnostics}
\end{figure*}

Action gating was also strongly task dependent. Across the four no-intervention diagnostics without matched corrective controls, aggregate success was only 20.0--23.8\% (Fig.~\ref{fig:action_gating_diagnostics}). Success reached 80--85\% when the \emph{Counting} goal was already satisfied and 0\% for both \emph{Object State} and \emph{Ordinal Position} across all three policies tested. This contrast shows that the policies can remain stationary but do not reliably condition that decision on whether the instruction already holds or has a valid referent.

\begin{figure}[t]
    \centering
    \includegraphics[width=.95\columnwidth]
    {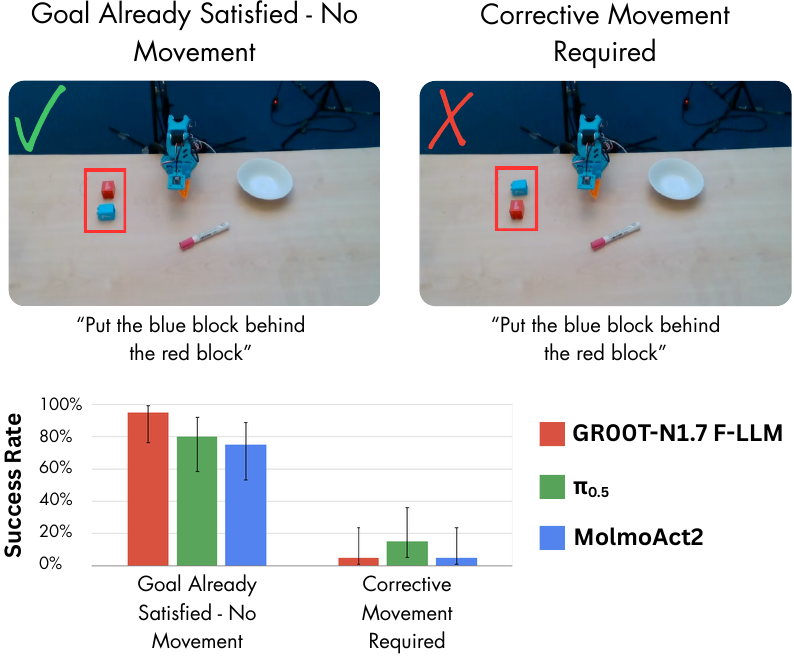}
    \caption{\textbf{Relational Placement action-gating control.} The same relation instruction is evaluated when the requested relation already holds and when corrective movement is required. Success is high in the no-movement condition but falls sharply when the policy must correct the relation, indicating that remaining stationary does not necessarily reflect reliable recognition that the goal is satisfied. Each condition contains 20 rollouts per policy; whiskers show two-sided Wilson 95\% confidence intervals.}
    \label{fig:no_movement_relational_placement}
\end{figure}

We analyze \emph{Relational Placement} separately because its no-movement scenes are paired with corresponding scenes requiring corrective movement (Fig.~\ref{fig:no_movement_relational_placement}). Success was 75--95\% when the requested relation already held but only 5--15\% when correction was required. The reversed performance gap suggests that the high no-movement success primarily reflects a task-specific tendency to remain stationary when the target and reference objects are nearby, rather than reliable recognition of whether intervention is required.

\subsection{Analysis of Failures by Task}

\begin{figure*}[t]
    \centering
    \includegraphics[width=\textwidth]{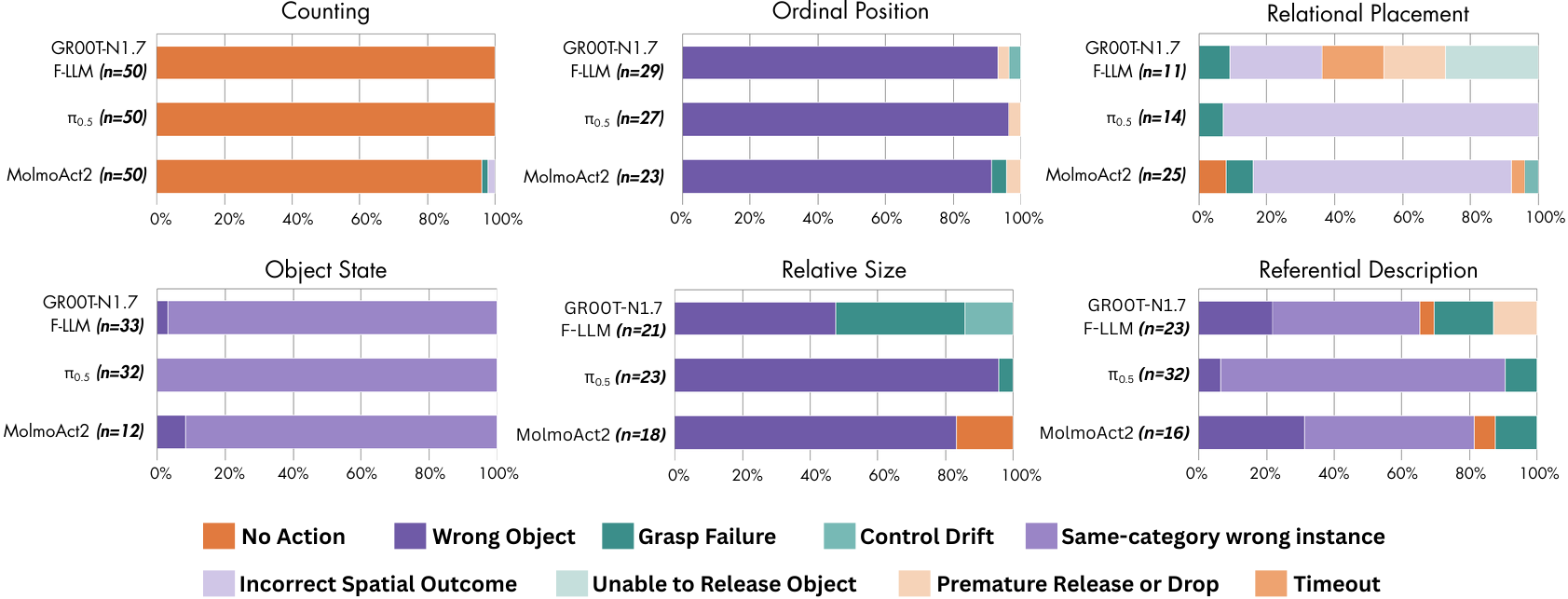}
    \caption{\textbf{Primary failure-mode composition for OOD spatial configurations.} Bars are normalized within each task--model pair, and \(n\) denotes the number of failures. Wrong-instance errors are separated from other wrong-target errors only for \textit{Object State} and \textit{Referential Description}.}
    \label{fig:failures_by_task_spatial_ood}
\end{figure*}

For each failed rollout, we recorded its behavior and assigned a mutually exclusive primary failure mode from the taxonomy in Fig.~\ref{fig:failures_by_task_spatial_ood}. When multiple problems occurred, we assigned the mode judged most directly responsible for failure. OOD patterns are strongly task dependent: no action accounts for 148/150 \textit{Counting} failures, wrong object selection for 74/79 \textit{Ordinal Position} failures, and same-category, wrong-instance selection for 75/77 \textit{Object State} failures. In contrast, most \textit{Relational Placement} failures from $\pi_{0.5}$ and MolmoAct2 involve incorrect spatial outcomes after approaching the intended object (32/39), while GR00T-N1.7 F-LLM exhibits a more heterogeneous mixture. These results highlight different bottlenecks across tasks and policies rather than a single failure mode that explains performance across the benchmark.

\section{Discussion}

The matched-scene results separate two capabilities that aggregate task success normally conflates. The high direct-manipulation success shows that the evaluated policies can often perceive, grasp, and move the relevant objects in the tested scenes. Their substantially lower success when the same motion must be selected from a spatial description therefore cannot be explained by manipulation difficulty alone. This does not imply that every OOD failure is purely linguistic: changing an instruction can alter attention, target selection, and the generated trajectory. It does show that successful motor execution under an explicit target name is not sufficient evidence of robust spatial grounding.

The withheld-cell split further indicates that exposure to each component of a behavior does not guarantee that those components will be recombined appropriately. Objects, spatial concepts, and required motions are individually represented during fine-tuning, yet performance decreases when their pairing is absent from spatially grounded instructions. The complete failure on OOD \textit{Counting} is the clearest instance: the requested counts and add/remove transitions are familiar, but none of the three policies uses the withheld count-conditioned combinations successfully. More generally, the task-level failure analysis distinguishes failures to initiate action, failures to select a target, and failures to realize a requested relation after approaching the intended object.

The additional diagnostics reveal two further dependencies. First, the cross-task-object decrease suggests that spatial behavior remains coupled to the object context in which it was demonstrated, even when the transferred object is familiar. Second, action gating is not a general ability shared across task formulations. Policies frequently remain still when a counting goal already holds, yet act despite invalid \textit{Object State} and \textit{Ordinal Position} instructions. In \textit{Relational Placement}, remaining stationary is highly successful when no movement is required but is also common when correction is necessary. Consequently, no-action success must be interpreted together with matched movement-required controls rather than as standalone evidence that the policy recognized the scene state.

Taken together, these findings argue against treating spatial grounding as a single scalar capability. Different tasks expose distinct bottlenecks, and policies with similar aggregate scores can fail for different reasons. Spa-Bench is therefore intended as a diagnostic suite: its value lies not only in ranking systems, but in identifying whether progress arises from better concept recombination, target selection, relation execution, linguistic robustness, or action gating.

\section{Limitations}

Spa-Bench deliberately fixes the robot embodiment, workspace, object inventory, receptacle, and manipulation family. This enables targeted attribution but limits external validity: scenes contain at most five candidate objects, resets are not pixel-identical, and the no-intervention conditions are zero-shot. The benchmark also tests one tabletop morphology and a restricted vocabulary of states, relations, counts, and ordinal descriptions. Direct-manipulation controls reduce motor execution as an alternative explanation but do not make the two language conditions identical, and the standard- versus cross-task-object comparison is descriptive rather than based on object-swapped scene pairs. Finally, model-specific preprocessing and adaptation choices mean that the results characterize complete policy systems rather than individual architectural components.

\section{Conclusion}

We introduced Spa-Bench to evaluate spatial grounding across six real-robot tasks. Across the three fully evaluated VLAs, success decreased from ID to OOD spatial configurations. On matched scenes, performance recovered when the target was named directly, and this advantage persisted under paraphrasing. Cross-task objects revealed sensitivity to learned object associations, while action-gating and failure annotations exposed task-specific bottlenecks in deciding whether to act, selecting a target, and realizing a spatial outcome. These findings show that successful manipulation does not necessarily imply spatial understanding. Spa-Bench's \emph{withheld-cell} design provides a controlled testbed for measuring whether VLAs select the correct behavior from the spatial information expressed in an instruction: a capability essential for reliable autonomy.

\clearpage
\appendices
\begin{figure*}[t]
    \centering
    \includegraphics[width=\textwidth]{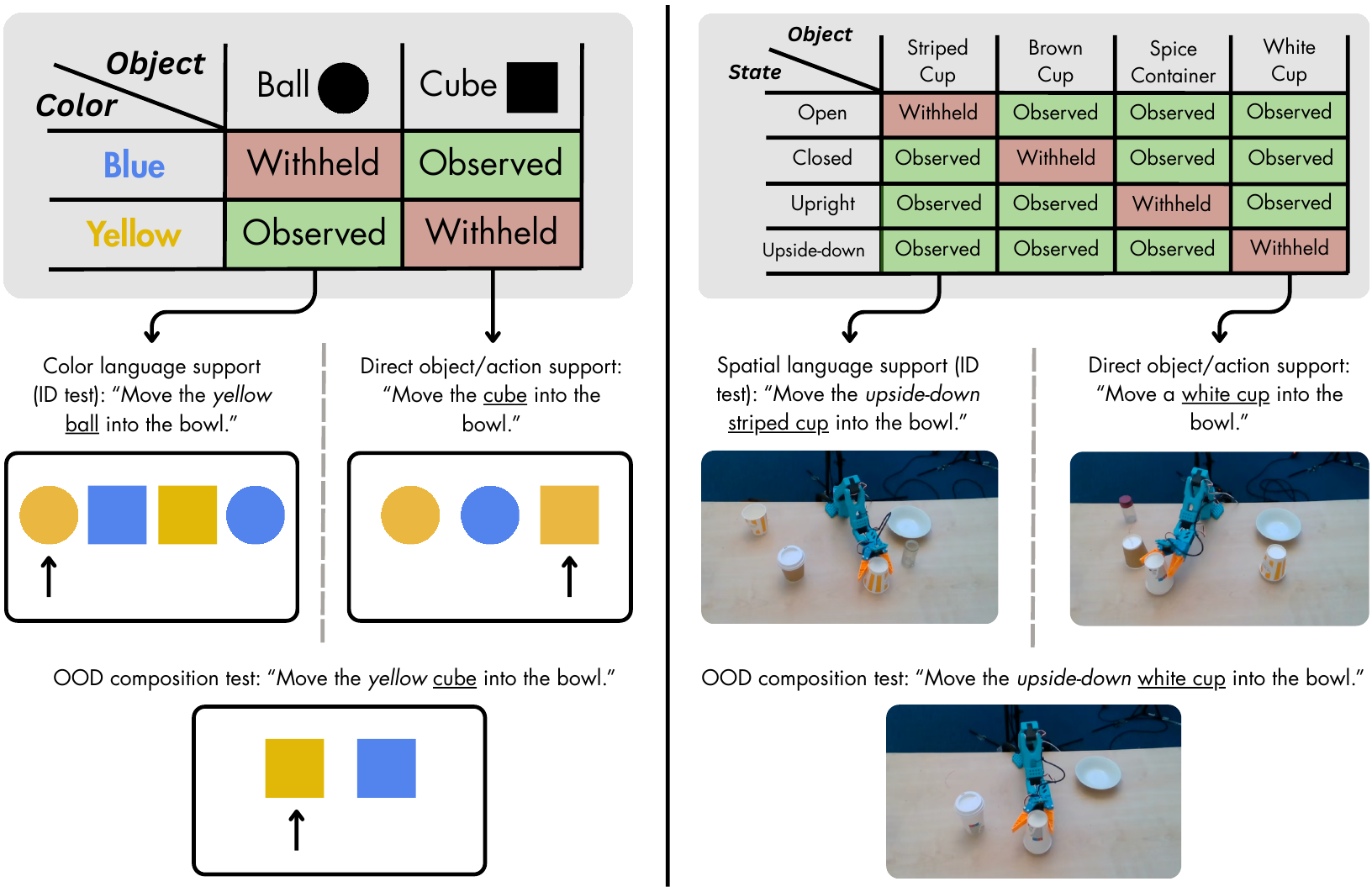}
    \caption{\textbf{Withheld-cell construction with preserved training support.} Left: a toy color--object example. Color language training supports the observed pairing \textit{yellow ball}, while a direct instruction demonstrates manipulation of the evaluation object without naming its color. The pairing \textit{yellow cube} is withheld from spatially grounded training and reserved for OOD evaluation. Right: the corresponding construction for \textit{Object State}. The concept \textit{upside-down} appears with the striped cup, and the white cup is manipulated under a direct instruction, but the pairing \textit{upside-down white cup} is withheld from spatially grounded training. OOD evaluation requires selecting the upside-down white cup rather than an upright white cup in the same scene.}
    \label{fig:withheld_cell_support_examples}
\end{figure*}

\section{Complete Withheld-Cell Design}
\label{app:complete_matrices}

Figure~\ref{fig:withheld_cell_support_examples} gives a simplified illustration of the withheld-cell construction and makes explicit how direct instructions preserve familiarity with the evaluation object and required manipulation. The same principle is then applied to each Spa-Bench task using task-specific spatial concepts and arguments.

\begin{figure*}[t]
    \centering
    \includegraphics[width=\textwidth]{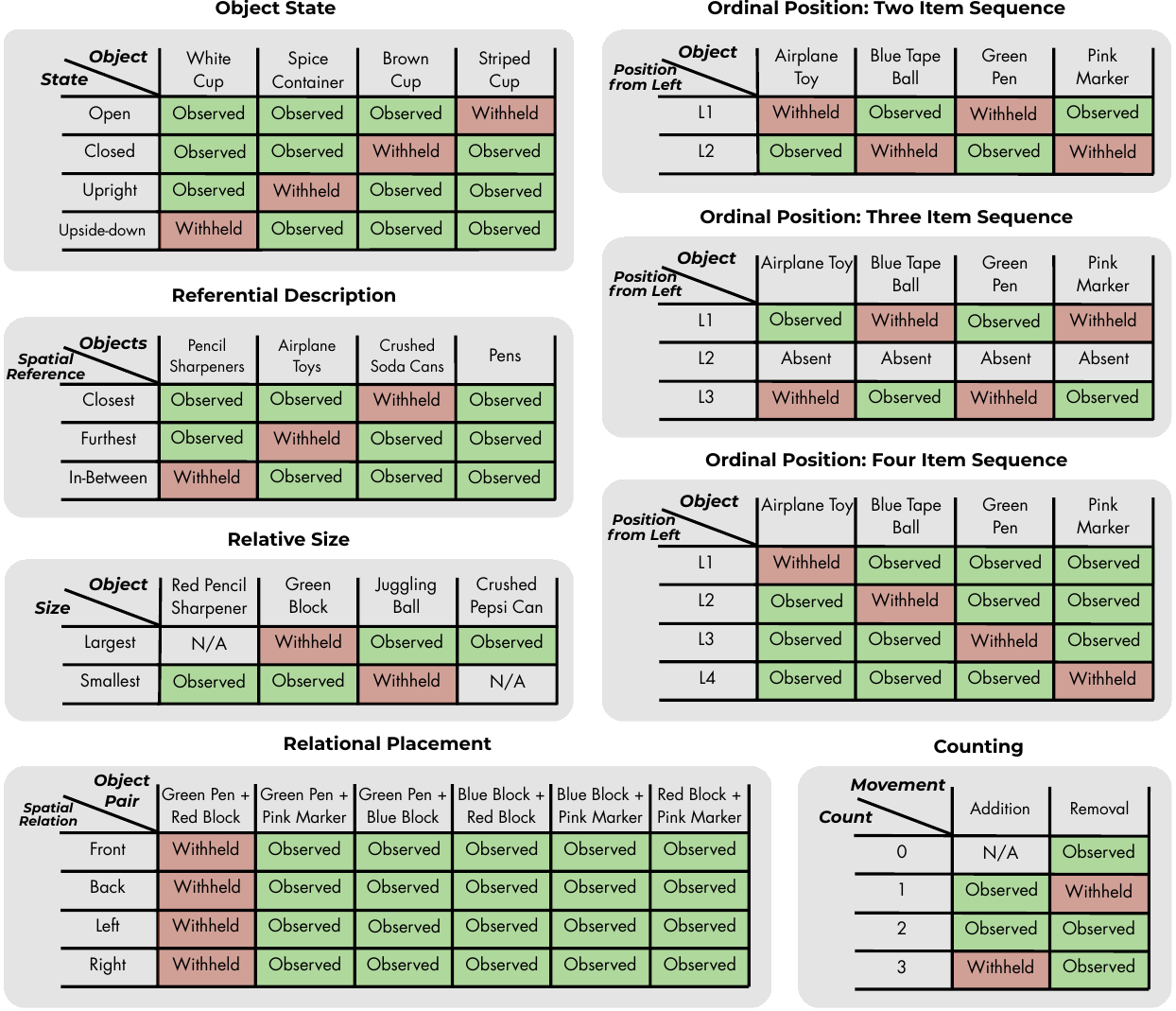}
    \caption{\textbf{Complete Spa-Bench concept--argument matrices.} Observed cells receive concept-conditioned instructions during fine-tuning; withheld cells are reserved for OOD spatial evaluation. Grey cells denote combinations that are absent or not applicable. The three-object \textit{Ordinal Position} matrix marks the middle position as absent because those training sequences target only the leftmost and rightmost positions.}
    \label{fig:complete_concept_argument_matrices}
\end{figure*}

Figure~\ref{fig:complete_concept_argument_matrices} summarizes the training support used to construct the ID and OOD evaluations. Green cells are explicitly paired with the corresponding spatial concept during fine-tuning. Red cells are withheld from spatially grounded instructions and reserved for OOD evaluation. Grey cells are absent or not applicable. In every withheld cell, the constituent object, spatial concept, and required physical behavior remain represented elsewhere in the fine-tuning data.

The task-specific constructions are as follows:

\begin{itemize}
    \item \textbf{Object State.} Every state word and object category appears during fine-tuning, but one state is withheld for each category: \emph{upside-down} for the white cup, \emph{open} for the striped cup, \emph{closed} for the brown cup, and \emph{upright} for the spice container. Explicit state demonstrations contain two objects from the same category, whereas state-masked direct demonstrations use a category-only instruction.

    \item \textbf{Relative Size.} The concepts \emph{smallest} and \emph{largest} appear with several objects. The green block is withheld from \emph{largest} instructions and the juggling ball from \emph{smallest} instructions, while both remain present in direct-object demonstrations.

    \item \textbf{Referential Description.} The concepts \emph{closest}, \emph{furthest}, and \emph{in-between} are represented across multiple candidate categories. The held-out combinations are crushed soda can--closest, airplane toy--furthest, and pencil sharpener--in-between. Pens form a fully crossed anchor category and receive all three relations.

    \item \textbf{Relational Placement.} The concepts \emph{left}, \emph{right}, \emph{front}, and \emph{behind} appear with multiple ordered object pairs. Directional instructions involving the green-pen--red-block pair are withheld. The same objects and trajectories remain represented under non-directional instructions such as ``Place the green pen next to the red block.''

    \item \textbf{Ordinal Position.} The training set contains left- and right-based descriptions, target identities, and sequence lengths of two, three, and four. Selected combinations of sequence length, physical position, and target identity are withheld. Three-object training sequences target only the leftmost and rightmost positions; consequently, the middle position is absent rather than treated as a controlled withheld cell.

    \item \textbf{Counting.} Goal counts of zero, one, two, and three and both addition and removal behaviors are represented. The withheld cells are a goal of one reached through a $2\rightarrow1$ removal and a goal of three reached through a $2\rightarrow3$ addition.
\end{itemize}

\section{Evaluation Conditions, Scene Matching, and Scoring}
\label{app:evaluation_details}

A complete Spa-Bench evaluation contains 900 physical rollouts. Each condition answers a separate diagnostic question and is not pooled into a single benchmark score. Table~\ref{tab:evaluation_conditions_appendix} lists the per-task allocation.

\begin{table*}[t]
\centering
\footnotesize
\caption{Spa-Bench evaluation conditions. Counts are per task unless otherwise noted.}
\label{tab:evaluation_conditions_appendix}
\renewcommand{\arraystretch}{1.12}
\begin{tabularx}{\textwidth}{p{0.22\textwidth}cX}
\toprule
\textbf{Condition} & \textbf{Trials} & \textbf{Diagnostic question} \\
\midrule
ID spatial configuration & 20 & Can the policy perform a concept--argument combination explicitly represented during fine-tuning? \\
OOD spatial configuration & 50 & Can the policy recombine individually supported concepts, arguments, and physical behaviors in a withheld configuration? \\
Matched direct manipulation & 20 & Can the policy execute the corresponding object motion in the same nominal scene when task-specific spatial inference is removed? \\
Prompt paraphrasing & 20 & Does performance change when 10 OOD spatial and 10 direct-manipulation instructions are rephrased? \\
Cross-task object & 20 & Does an ID spatial behavior transfer to an object associated with a different Spa-Bench task? \\
No intervention & 20 & Can the policy refrain from acting when the requested goal already holds or no valid referent exists? Not used for \textit{Relative Size}. \\
Corrective movement & 20 & For \textit{Relational Placement} only, will the policy act when nearby objects do not yet satisfy the requested relation? \\
\bottomrule
\end{tabularx}
\end{table*}

Every task contains 20 ID and 50 OOD spatial-configuration trials. Twenty OOD scenes per task are repeated under matched direct-manipulation instructions, and 10 of those scenes per task are additionally repeated with paraphrases of both instruction types. Cross-task-object scene specifications are held constant across policies within that condition, but they are not one-to-one replicas of the standard-object ID scenes with only the target identity changed. Their aggregate difference from the standard-object ID condition is therefore descriptive.

Five tasks contain 20 no-intervention trials in which the correct response is to remain stationary. \textit{Relative Size} excludes this condition because every comparison set contains a unique smallest and largest object. \textit{Relational Placement} additionally contains 20 movement-required controls. Consequently, \textit{Relational Placement} contains 170 trials, \textit{Relative Size} contains 130, and each remaining task contains 150, giving 900 rollouts per fully evaluated policy.

Corresponding policy rollouts are manually reset to the same scene specification, including object identities and nominal locations. Reference images guide each reset, but the resulting scenes are not pixel-identical. Each rollout receives a binary task-success label using the criteria in Table~\ref{tab:task_overview}; partial credit is not assigned. All policies except VLA-0 receive a 60-second time limit. VLA-0 receives 90 seconds because its autoregressive action generation is slower. A no-intervention trial succeeds only when the policy avoids task-directed arm or gripper movement. No human correction occurs after the rollout timer and scoring begin.

\section{Model Adaptation and Deployment Details}
\label{app:model_details}

\subsection{Inputs and Frame Processing}

Every policy receives the six-dimensional robot state and two $480\times640$ RGB observations: a fixed middle-camera view facing the robot and a wrist-mounted view. Each image is converted to the policy's native input resolution during preprocessing. The released dataset also contains left, right, and overhead observations, but these views are not supplied to the evaluated policies.

The episode identities and semantic content are shared across policies, but frame-level preprocessing is not identical. MolmoAct2, $\pi_{0.5}$, and GR00T-N1.7 F-FT use the complete 612,733-frame dataset. GR00T-N1.7 F-LLM and VLA-0 use a motion-trimmed derivative containing 570,386 frames. Motion onset is defined as the first of five consecutive frames whose maximum joint displacement from the initial five-frame baseline exceeds 2.0 motor-position units. Five frames preceding the detected onset are retained. This removes 42,333 idle-prefix frames from 1,176 episodes. A further 14 frames repair terminal camera--state alignment in two episodes, yielding 42,347 removed frames, or 6.91\% of the source dataset. Prompts, actions, states, and episode identities are otherwise preserved.

Motion trimming was introduced after preliminary runs showed that shorter-horizon policies could remain at the static initial configuration rather than initiate the task. It is therefore a model-specific adaptation choice rather than a controlled architectural variable.

\subsection{Optimization and Model-Specific Adaptation}

All five policies are trained on four NVIDIA GH200 Grace Hopper Superchips with a per-device batch size of 24 and a global batch size of 96. The checkpoint at the end of epoch 12 is used for every reported evaluation. Table~\ref{tab:training_configuration_appendix} summarizes the resolved adaptation settings.

\begin{table*}[t]
\centering
\scriptsize
\setlength{\tabcolsep}{2.5pt}
\caption{Resolved model-adaptation settings. Full and trimmed denote the 612,733-frame and 570,386-frame dataset variants, respectively; $H$ is the predicted action horizon.}
\label{tab:training_configuration_appendix}
\begin{tabular}{p{0.14\textwidth}p{0.14\textwidth}p{0.10\textwidth}p{0.04\textwidth}p{0.29\textwidth}p{0.20\textwidth}}
\toprule
\textbf{Policy} & \textbf{Initialization} & \textbf{Data / updates} & \textbf{$H$} & \textbf{Updated components} & \textbf{Action/state processing} \\
\midrule
$\pi_{0.5}$ & \texttt{lerobot/pi05\_base} & Full / 76,596 & 50 & Vision-language backbone and action expert & Quantile normalization \\
MolmoAct2 & \texttt{allenai/MolmoAct2} & Full / 76,596 & 30 & VLM and action expert; token embeddings frozen & Min--max normalization \\
VLA-0 & Qwen2.5-VL-3B-Instruct & Trimmed / 77,246 & 8 & Full Qwen backbone; no LoRA or QLoRA & Min--max scaling; 1,000 text bins \\
GR00T-N1.7 F-LLM & GR00T-N1.7-3B & Trimmed / 71,304 & 16 & Visual encoder, multimodal projector, VLLN, and diffusion-action model; LLM frozen & GR00T-native processing \\
GR00T-N1.7 F-FT & GR00T-N1.7-3B & Full / 76,596 & 16 & Language, visual, multimodal, and diffusion-action components & GR00T-native processing \\
\bottomrule
\end{tabular}
\end{table*}

All policies use AdamW. MolmoAct2 uses a base learning rate of $10^{-5}$, zero weight decay, 200 warm-up updates, and cosine decay; its vision transformer and connector use $5\times10^{-6}$, while its action expert uses $5\times10^{-5}$. The $\pi_{0.5}$ configuration uses a learning rate of $2.5\times10^{-5}$, weight decay of 0.01, 1,000 warm-up updates, and cosine decay. VLA-0 uses a learning rate of $5\times10^{-6}$, weight decay of 0.01, and cosine annealing. Its preprocessing resizes both camera observations to $224\times224$ pixels and applies random cropping and color jitter during training. Both GR00T-N1.7 configurations use weight decay $10^{-5}$ and a 5\% warm-up followed by cosine decay; F-LLM uses a learning rate of $10^{-4}$ and F-FT uses $10^{-5}$.

\subsection{Deployment-Time Processing}

During unscored preliminary deployment, both GR00T-N1.7 configurations exhibited high-frequency action jitter that interfered with grasp execution. Following the deployment procedure described by ANR Robot \cite{anr_gr00t_experiments}, the same causal exponential moving-average filter is applied to every non-gripper action dimension for both policies:

\begin{equation}
    \widetilde{a}_{t,j}=0.25a_{t,j}+0.75\widetilde{a}_{t-1,j},
    \label{eq:groot_ema_appendix}
\end{equation}

where $a_{t,j}$ is the predicted target for action dimension $j$ and $\widetilde{a}_{t,j}$ is the executed target. The filter is initialized from the robot's measured configuration, persists across action-chunk boundaries, and resets between rollouts. The gripper command is not filtered so that grasp closure is not delayed.

GR00T-N1.7 F-FT retained the static initial prefixes during training and frequently remained stationary at the nominal starting configuration during preliminary, unscored deployment. For this policy only, a fixed small upward displacement is applied immediately before autonomous execution. The displacement occurs before the rollout timer and scoring begin, and no human correction is supplied during the scored rollout. GR00T-N1.7 F-LLM and VLA-0 do not receive this assist.

The displacement was not logged directly, so its magnitude was reconstructed retrospectively from the first five frames of the 120 GR00T-N1.7 F-FT rollouts and their corresponding GR00T-N1.7 F-LLM trials. Forward kinematics using the official SO-101 URDF produced a mean starting-pose separation of 10.09~mm, including a mean upward component of 7.63~mm. The mean paired maximum absolute arm-joint difference was $6.40^{\circ}$. Relative to the nominal initialization pose, the mean end-effector deviations were 9.27~mm for GR00T-N1.7 F-FT and 3.09~mm for GR00T-N1.7 F-LLM. These are retrospective estimates and include ordinary reset, calibration, and measurement variation; the intentional displacement for F-LLM was zero.

These model-specific choices remain fixed across all evaluated task families and conditions for the affected policy. Comparisons therefore concern complete adapted policy systems---including pretrained weights, action horizon, normalization, preprocessing, and deployment controller---rather than isolating any one architectural component.

\clearpage
\onecolumn

\section{Complete Numerical Results}
\label{app:numerical_results}

All intervals in this section are two-sided Wilson 95\% confidence intervals. Values are reported as successes/trials followed by the success rate and confidence interval expressed as percentages.

\begin{table}[htbp]
\centering
\scriptsize
\caption{Aggregate results for the three fully evaluated policies. Matched OOD and matched direct manipulation use the same 120 nominal scene specifications.}
\label{tab:aggregate_results_appendix}
\renewcommand{\arraystretch}{1.18}
\begin{tabular}{lcccc}
\toprule
\textbf{Policy} & \textbf{ID} & \textbf{All OOD} & \textbf{Matched OOD} & \textbf{Matched direct} \\
\midrule
GR00T-N1.7 F-LLM & 81/120 (67.5\%, 58.7--75.2) & 133/300 (44.3\%, 38.8--50.0) & 50/120 (41.7\%, 33.2--50.6) & 98/120 (81.7\%, 73.8--87.6) \\
$\pi_{0.5}$ & 82/120 (68.3\%, 59.6--76.0) & 122/300 (40.7\%, 35.3--46.3) & 54/120 (45.0\%, 36.4--53.9) & 100/120 (83.3\%, 75.7--88.9) \\
MolmoAct2 & 88/120 (73.3\%, 64.8--80.4) & 156/300 (52.0\%, 46.4--57.6) & 73/120 (60.8\%, 51.9--69.1) & 107/120 (89.2\%, 82.3--93.6) \\
\bottomrule
\end{tabular}
\end{table}

\begin{table}[htbp]
\centering
\scriptsize
\caption{Task-level ID spatial-grounding results. Each cell reports successes/20, success rate, and Wilson 95\% confidence interval.}
\label{tab:id_task_results_appendix}
\renewcommand{\arraystretch}{1.18}
\begin{tabular}{lccc}
\toprule
\textbf{Task} & \textbf{GR00T-N1.7 F-LLM} & \textbf{$\pi_{0.5}$} & \textbf{MolmoAct2} \\
\midrule
Counting & 12/20; 60.0\% (38.7--78.1) & 11/20; 55.0\% (34.2--74.2) & 12/20; 60.0\% (38.7--78.1) \\
Ordinal Position & 9/20; 45.0\% (25.8--65.8) & 7/20; 35.0\% (18.1--56.7) & 10/20; 50.0\% (29.9--70.1) \\
Relational Placement & 14/20; 70.0\% (48.1--85.5) & 13/20; 65.0\% (43.3--81.9) & 16/20; 80.0\% (58.4--91.9) \\
Object State & 17/20; 85.0\% (64.0--94.8) & 16/20; 80.0\% (58.4--91.9) & 18/20; 90.0\% (69.9--97.2) \\
Relative Size & 17/20; 85.0\% (64.0--94.8) & 19/20; 95.0\% (76.4--99.1) & 17/20; 85.0\% (64.0--94.8) \\
Referential Description & 12/20; 60.0\% (38.7--78.1) & 16/20; 80.0\% (58.4--91.9) & 15/20; 75.0\% (53.1--88.8) \\
\bottomrule
\end{tabular}
\end{table}

\begin{table}[htbp]
\centering
\scriptsize
\caption{Task-level OOD spatial-grounding results. Each cell reports successes/50, success rate, and Wilson 95\% confidence interval.}
\label{tab:ood_task_results_appendix}
\renewcommand{\arraystretch}{1.18}
\begin{tabular}{lccc}
\toprule
\textbf{Task} & \textbf{GR00T-N1.7 F-LLM} & \textbf{$\pi_{0.5}$} & \textbf{MolmoAct2} \\
\midrule
Counting & 0/50; 0.0\% (0.0--7.1) & 0/50; 0.0\% (0.0--7.1) & 0/50; 0.0\% (0.0--7.1) \\
Ordinal Position & 21/50; 42.0\% (29.4--55.8) & 23/50; 46.0\% (33.0--59.6) & 27/50; 54.0\% (40.4--67.0) \\
Relational Placement & 39/50; 78.0\% (64.8--87.2) & 36/50; 72.0\% (58.3--82.5) & 25/50; 50.0\% (36.6--63.4) \\
Object State & 17/50; 34.0\% (22.4--47.8) & 18/50; 36.0\% (24.1--49.9) & 38/50; 76.0\% (62.6--85.7) \\
Relative Size & 29/50; 58.0\% (44.2--70.6) & 27/50; 54.0\% (40.4--67.0) & 32/50; 64.0\% (50.1--75.9) \\
Referential Description & 27/50; 54.0\% (40.4--67.0) & 18/50; 36.0\% (24.1--49.9) & 34/50; 68.0\% (54.2--79.2) \\
\bottomrule
\end{tabular}
\end{table}

\begin{table}[htbp]
\centering
\scriptsize
\caption{Task-level matched direct-manipulation results. Each cell reports successes/20, success rate, and Wilson 95\% confidence interval.}
\label{tab:direct_task_results_appendix}
\renewcommand{\arraystretch}{1.18}
\begin{tabular}{lccc}
\toprule
\textbf{Task} & \textbf{GR00T-N1.7 F-LLM} & \textbf{$\pi_{0.5}$} & \textbf{MolmoAct2} \\
\midrule
Counting & 12/20; 60.0\% (38.7--78.1) & 11/20; 55.0\% (34.2--74.2) & 16/20; 80.0\% (58.4--91.9) \\
Ordinal Position & 19/20; 95.0\% (76.4--99.1) & 20/20; 100\% (83.9--100) & 17/20; 85.0\% (64.0--94.8) \\
Relational Placement & 17/20; 85.0\% (64.0--94.8) & 18/20; 90.0\% (69.9--97.2) & 17/20; 85.0\% (64.0--94.8) \\
Object State & 19/20; 95.0\% (76.4--99.1) & 18/20; 90.0\% (69.9--97.2) & 20/20; 100\% (83.9--100) \\
Relative Size & 19/20; 95.0\% (76.4--99.1) & 20/20; 100\% (83.9--100) & 20/20; 100\% (83.9--100) \\
Referential Description & 12/20; 60.0\% (38.7--78.1) & 13/20; 65.0\% (43.3--81.9) & 17/20; 85.0\% (64.0--94.8) \\
\bottomrule
\end{tabular}
\end{table}

\begin{table}[htbp]
\centering
\scriptsize
\caption{Prompt-paraphrase results on the corresponding 60-scene subsets.}
\label{tab:paraphrase_results_appendix}
\renewcommand{\arraystretch}{1.18}
\begin{tabular}{lccc}
\toprule
\textbf{Condition} & \textbf{GR00T-N1.7 F-LLM} & \textbf{$\pi_{0.5}$} & \textbf{MolmoAct2} \\
\midrule
Original direct & 49/60; 81.7\% (70.1--89.4) & 52/60; 86.7\% (75.8--93.1) & 54/60; 90.0\% (79.9--95.3) \\
Paraphrased direct & 42/60; 70.0\% (57.5--80.1) & 52/60; 86.7\% (75.8--93.1) & 54/60; 90.0\% (79.9--95.3) \\
Original OOD spatial & 29/60; 48.3\% (36.2--60.7) & 29/60; 48.3\% (36.2--60.7) & 39/60; 65.0\% (52.4--75.8) \\
Paraphrased OOD spatial & 23/60; 38.3\% (27.1--51.0) & 25/60; 41.7\% (30.1--54.3) & 30/60; 50.0\% (37.7--62.3) \\
\bottomrule
\end{tabular}
\end{table}

\begin{table}[htbp]
\centering
\scriptsize
\caption{Cross-task-object and action-gating diagnostics. Cross-task rows contain 100 trials; non-relational task rows and relational-control rows contain 20 trials; the four-task aggregates contain 80 trials.}
\label{tab:additional_diagnostics_appendix}
\renewcommand{\arraystretch}{1.18}
\begin{tabular}{lccc}
\toprule
\textbf{Condition} & \textbf{GR00T-N1.7 F-LLM} & \textbf{$\pi_{0.5}$} & \textbf{MolmoAct2} \\
\midrule
Standard objects, five-task ID & 64/100; 64.0\% (54.2--72.7) & 66/100; 66.0\% (56.3--74.5) & 70/100; 70.0\% (60.4--78.1) \\
Cross-task objects, five-task ID & 55/100; 55.0\% (45.2--64.4) & 54/100; 54.0\% (44.3--63.4) & 53/100; 53.0\% (43.3--62.5) \\
\midrule
Counting: goal already satisfied & 17/20; 85.0\% (64.0--94.8) & 16/20; 80.0\% (58.4--91.9) & 16/20; 80.0\% (58.4--91.9) \\
Ordinal Position: invalid ordinal & 0/20; 0.0\% (0.0--16.1) & 0/20; 0.0\% (0.0--16.1) & 0/20; 0.0\% (0.0--16.1) \\
Object State: no valid target & 0/20; 0.0\% (0.0--16.1) & 0/20; 0.0\% (0.0--16.1) & 0/20; 0.0\% (0.0--16.1) \\
Referential Description: no candidate & 0/20; 0.0\% (0.0--16.1) & 0/20; 0.0\% (0.0--16.1) & 3/20; 15.0\% (5.2--36.0) \\
Four-task no-intervention aggregate & 17/80; 21.2\% (13.7--31.4) & 16/80; 20.0\% (12.7--30.0) & 19/80; 23.8\% (15.8--34.1) \\
\midrule
Relational goal already satisfied & 19/20; 95.0\% (76.4--99.1) & 16/20; 80.0\% (58.4--91.9) & 15/20; 75.0\% (53.1--88.8) \\
Relational corrective movement & 1/20; 5.0\% (0.9--23.6) & 3/20; 15.0\% (5.2--36.0) & 1/20; 5.0\% (0.9--23.6) \\
\bottomrule
\end{tabular}
\end{table}

\clearpage

\section{Extended Failure Analysis}
\label{app:extended_failures}

For each failed rollout, the evaluator recorded the observed behavior and apparent cause of failure. After evaluation, these notes were organized into the nine mutually exclusive primary modes in Table~\ref{tab:failure_taxonomy_appendix}. When multiple problems occurred, the mode judged most directly responsible for the unsuccessful outcome was used as the primary label; additional problems were retained as secondary annotations. These annotations remain separate from binary task scoring and do not affect the quantitative success rates.

\begin{table}[htbp]
\centering
\footnotesize
\caption{Primary qualitative failure-mode taxonomy.}
\label{tab:failure_taxonomy_appendix}
\renewcommand{\arraystretch}{1.12}
\begin{tabularx}{\textwidth}{p{0.27\textwidth}X}
\toprule
\textbf{Failure mode} & \textbf{Operational description} \\
\midrule
No action & The policy does not initiate task-directed arm or gripper motion when movement is required. \\
Wrong object & The policy approaches or manipulates an object outside the intended target category or otherwise selects the wrong target. \\
Correct category, wrong instance & The policy selects an object from the requested semantic category but chooses the instance with the wrong state or referential relation. Used only for \textit{Object State} and \textit{Referential Description}. \\
Grasp failure & The policy selects the intended target but fails to establish or maintain a usable grasp before transport. \\
Incorrect spatial outcome & The intended object is approached or manipulated, but the final count, direction, relation, or placement is incorrect. \\
Premature release or drop & The policy releases the object before reaching the intended goal or unintentionally drops it during transport. \\
Unable to release object & The policy reaches or approaches the intended goal but does not open the gripper sufficiently to complete placement. \\
Control drift & The executed motion drifts away from the intended trajectory or target without a more specific preceding semantic error. \\
Timeout & The rollout begins a potentially relevant behavior but does not complete the task within the evaluation limit. \\
\bottomrule
\end{tabularx}
\end{table}

\begin{figure}[htbp]
    \centering
    \includegraphics[width=\textwidth]{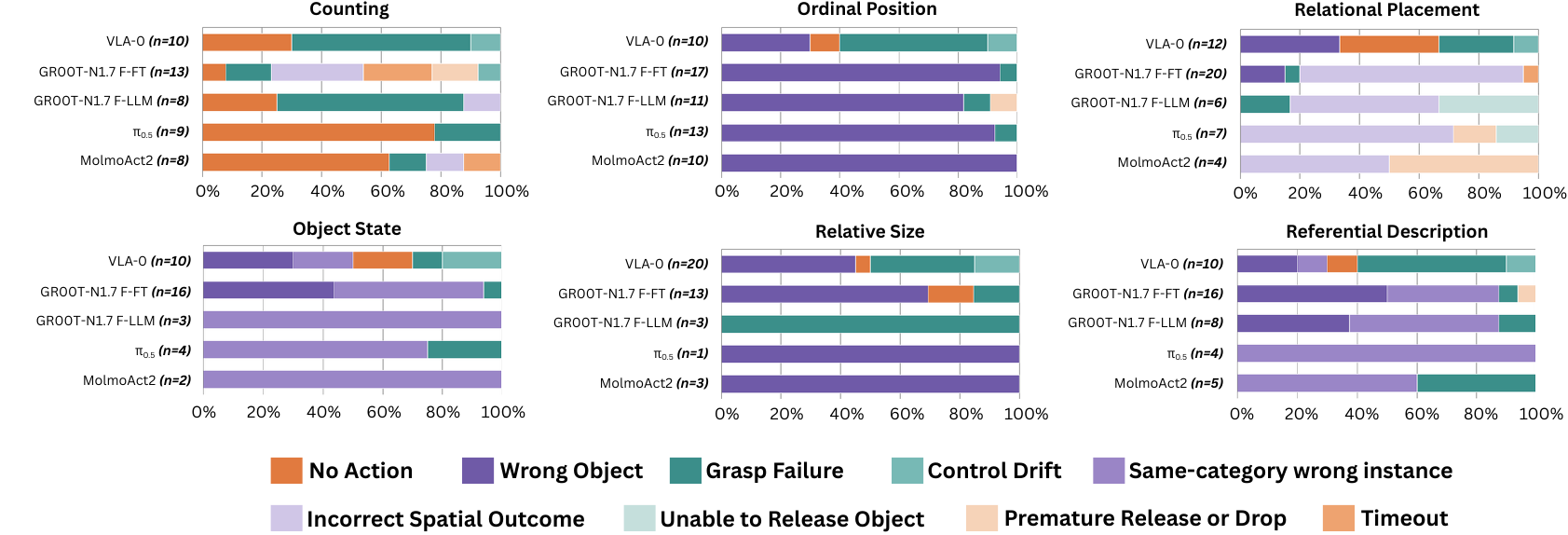}
    \caption{\textbf{Primary failure-mode composition for ID spatial configurations.} Bars are normalized within each task--policy pair, and $n$ denotes the number of failed rollouts. Results include all five policies; VLA-0 has fewer ID evaluation trials than the other policies (Fig.~\ref{fig:familiar_by_task}). Wrong-instance errors are separated from other wrong-target errors only for \textit{Object State} and \textit{Referential Description}. Some bars represent very few failures and are therefore descriptive.}
    \label{fig:failure_modes_id_appendix}
\end{figure}

Figure~\ref{fig:failure_modes_id_appendix} complements the OOD failure analysis in Fig.~\ref{fig:failures_by_task_spatial_ood} by showing ID failure-mode composition for all five policies, including the two diagnostic ID baselines, GR00T-N1.7 F-FT and VLA-0. The task-level ID success rates are reported in Fig.~\ref{fig:familiar_by_task}. These distributions describe difficulties that arise even for training-supported spatial combinations. Each bar is normalized by the number of failures for its task--policy pair, so the proportions do not indicate overall task success or failure rates. Pairs with few failures should be interpreted descriptively.

\end{document}